\documentclass[a4paper]{article}
\usepackage{INTERSPEECH2020}
\usepackage{xcolor}

\title{End-to-end Neural Diarization: From Transformer to Conformer}

\name{
Yi Chieh Liu$^{1}$\textsuperscript{\textsection}  \quad 
Eunjung Han$^{2}$\textsuperscript{\textsection}  \quad 
Chul Lee$^2$ \quad
Andreas Stolcke$^2$}
\address{
  $^1$Georgia Institute of Technology   \quad 
  $^2$Amazon Alexa Speech
}
\email{yliu3233@gatech.edu  \quad\quad  \{cehan,chulle,stolcke\}@amazon.com}

\begin{document}

\maketitle
\begin{abstract}
We propose a new end-to-end neural diarization (EEND) system that is based on Conformer, a recently proposed neural architecture that combines convolutional mappings and Transformer to model both local and global dependencies in speech.
We first show that data augmentation and convolutional subsampling layers enhance the original self-attentive EEND in the Transformer-based EEND, and then Conformer gives an additional gain over the Transformer-based EEND.
However, we notice that the Conformer-based EEND does not generalize as well from simulated to real conversation data as the Transformer-based model. This leads us to quantify the mismatch between simulated data and real speaker behavior in terms of temporal statistics reflecting turn-taking between speakers, and investigate its correlation with diarization error. By mixing simulated and real data in EEND training, we mitigate the mismatch further, with Conformer-based EEND achieving 
24\% error reduction over the baseline SA-EEND system, and 10\% improvement over the best augmented Transformer-based system, on two-speaker CALLHOME data.
\end{abstract}


\begingroup\renewcommand\thefootnote{\textsection}
\footnotetext{Equal contribution. The first author was an intern at Amazon.}

\section{Introduction}

 Speaker diarization is the process of partitioning an audio recording into homogeneous segments according to the speaker’s identity. It enables various speech applications to understand natural conversations in different scenarios, such as meetings or single-channel telephone conversations. In multi-speaker ASR, speaker-attributed transcripts are generated via speaker turn-taking analysis from a diarization system \cite{boeddeker2018front, kanda2018hitachi, kanda2019acoustic}. In target-speaker ASR, speaker embeddings extracted by diarization are employed to improve accuracy \cite{kanda2019simultaneous}.

Clustering of speaker embeddings from short speech segments is a commonly used technique in conventional speaker diarization systems \cite{shum2013unsupervised, sell2014speaker, garcia2017speaker}. 
While classical clustering-based diarization systems achieve an overall good performance, these have several shortcomings. First, they rely on multiple modules (VAD, embedding extractor, etc.) that need to be trained separately. This means that clustering-based diarization systems require a joint calibration process across different modules, introducing an additional complexity in training. Second, clustering of speaker embeddings is not a natural way of properly capturing speech overlaps in spite of some recent work to handle clusters with simultaneously active speakers \cite{bullock2020overlap}.
The self-attentive end-to-end neural diarization (SA-EEND) \cite{fujita2019end} is one of the state-of-the-art approaches aiming to model the joint speech activity of multiple speakers. SA-EEND integrates voice activity and overlap detection with speaker tracking in end-to-end fashion.  Moreover, it directly minimizes diarization errors and has demonstrated excellent diarization accuracy on two-speaker telephone conversation data sets.  

Recently, Transformer or Conformer based transducers \cite{zhang2020transformer,Gulati2020} have demonstrated a significant accuracy improvement in end-to-end ASR systems, compared to previous recurrent neural network transducers (RNN-T) \cite{he2019streaming}. Inspired by these advances, we propose a novel Conformer-based EEND system to investigate if we can further improve the previous state-of-the-art SA-EEND system \cite{fujita2019end}.
We show that our Transformer-based EEND benefits from data augmentation and convolutional subsampling over the original SA-EEND, and that Conformer then gives a further gain over the Transformer-based EEND. More importantly, as EEND requires training on simulated data, the degree of mismatch between training and test data becomes a key for the performance of such models, as has been investigated for transfer learning \cite{moore2010intelligent, cui2018large}.

Our main contributions are: (1) We propose a novel Conformer-based EEND model that consists of Conformer blocks, data augmentation layer, and convolutional subsampling layer, showing that such a model outperforms previous state-of-the-art SA-EEND models based on Transformers.
(2) We show that data augmentation and convolutional subsampling layers play a critical role. When we enhance the original SA-EEND models with data augmentation and convolutional subsampling layers, we observe substantial accuracy improvements.
(3) While our Conformer-based EEND model is superior to Transformer-based EEND in overall performance, we show that our proposed model is less robust to the train/test mismatch 
that results from data simulation. We quantify the degree of mismatch with statistics related to turn-taking (overlap and non-speech durations), and combine simulated with real data to mitigate its effect, yielding overall best results.

\section{Transformer-based EEND}
In the original self-attentive EEND (SA-EEND) model \cite{fujita2019end}, speaker diarization is formulated as a multi-label classification problem. That is, given a $T$-length sequence of $F$-dimensional audio features, $X=[x_1, \dots, x_T], \ x_t \in R^F$, stacked frames ($X')$ are sub-sampled by a factor of ten (i.e. $T' = T/10$) before sending them to a Transformer encoder. The Transformer encoder computes $D$-dimensional diarization embeddings, $E = [e_1, \dots,  e_{T'}], \ e_t \in R^D$.  Frame-wise posterior probabilities of the joint speech activities for $S$ (fixed) speakers, $Z=  [z_1, \dots, z_{T'}], \ z_t \in R^S$, are estimated using a fully connected layer with element-wise sigmoid function.
The system estimates a sequence of joint speech activities for $S$ speakers, $\hat{Y} = [\hat{y}_1, \dots,  \hat{y}_{T'}], \ \hat{y}_t \in R^S$. For a posterior probability $z_{t,s}$, speaker $s$ is active at time index $t$ ($\hat{y}_{t,s} = 1$) if it is above a certain threshold while it is inactive ($\hat{y}_{t,s} = 0$) otherwise. During training, a threshold of 0.5 is used. To deal with permutation ambiguity in speaker labeling, a  training scheme that considers all permutations of the reference speaker labels is employed.

We further enhance the original SA-EEND by adding a data augmentation layer, SpecAugment \cite{park2019specaugment}, which is followed by convolutional subsampling instead of subsampling after frame stacking. We call our enhanced model as Transformer-based EEND (TB-EEND).
\begin{equation} 
\label{eq2}
\begin{split}
X' &= \text{SpecAugment}(X) \\
X'' &= \text{ConvolutionalSubsampling}(X') \\
E &= \text{TransformerEncoder}(X'') \\
Z &= \sigma(\text{Linear}(E))
\end{split}
\end{equation}
The convolution subsampling layer captures relative positional information and local interactions between nearby features.

\section{Conformer-based EEND}

\begin{figure}[t]
\centering
\includegraphics[width=0.4\textwidth]{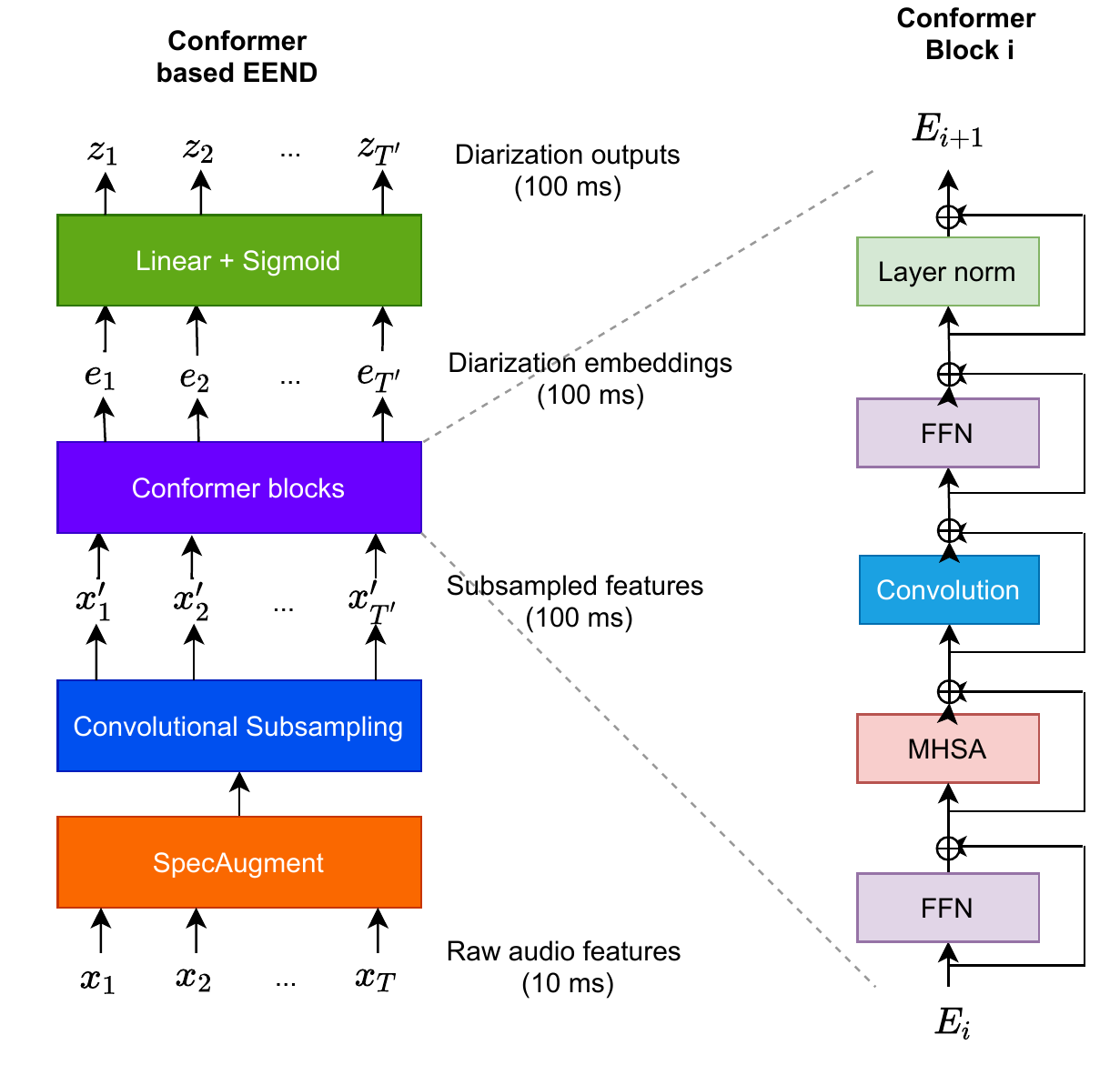}
\caption{Schematic Diagram of Conformer-based EEND.}
\label{fig:conformer}
\end{figure}

Recently, Conformer was proposed \cite{Gulati2020} as an ASR encoder architecture that combines local and global dependency modeling.
Since diarization relies both on local cues (e.g., at speaker changes) and long-range comparisons of speaker characteristics,
we propose a novel Conformer-based EEND (CB-EEND), as shown in Figure \ref{fig:conformer}.
The main idea is to replace the Transformer encoder in the equation (\ref{eq2}) of the Transformer-based EEND model with a Conformer encoder.
Each Conformer block, in turn, is composed of four stacked modules, including the first feed-forward network (FFN) module, a multi-head self-attention (MHSA) module, a convolution module, and the second FFN module. The split feed-forward modules are known as sandwich structures inspired by \cite{lu*2020understanding}. Given the input $E_i$ to the $i$-th Conformer block, the output $E_{i+1}$ is computed as follows:
\begin{equation} 
\begin{aligned}
 &\tilde{E}_i =E_i+\frac{1}{2}\text{FFN}(E_i)\\
 &\tilde{E}_i'  = \tilde{E}_i + \text{MHSA}( \tilde{E}_i)\\
 &\tilde{E}_i'' = \tilde{E}_i'+ \text{Convolution}(\tilde{E}_i')\\
 &E_{i+1}     = \text{LayerNorm}(\tilde{E}_i''+\frac{1}{2}\text{FFN}(\tilde{E}_i''))
 \end{aligned}
\end{equation}
In the later experimental section, we study the impact of different parameters in SpecAugment, convolution subsampling and Conformer blocks as ablation studies.

\section{Conversational Similarity}

The inherent difficulty of the diarization task is greatly influenced by the turn-taking behavior of speakers, in particular, how much speakers are separated by non-speech or overlap in their speech.
Since EEND systems implicitly learn the statistics of turn-taking, it is important to ask to what extent they affect diarization performance, and how well the simulated data used in EEND training matches real (target domain) conversations.
As a first step in characterizing the relevant properties of speech data, we compare different data sets used in our experiments along those two dimensions: durations of {\em speech overlap} and durations of {\em silence (non-speech)} regions.
Following Cui et al.~\cite{cui2018large}, we quantify the mismatch between data sets by computing the Earth Mover’s distance (EMD) \cite{rachev1985monge, rubner2000earth} between the respective distributions.

Let $S$ and $T$ denote the cumulative distribution functions (CDF) that represents the variable interest in source and target data, respectively.
Let $s_i$ and $t_i$ denote the $i$-th elements in the source ($S$) and target ($T$) CDFs, respectively.
The EMD describes how much and how far probability mass has to be moved to turn the $S$ distribution into $T$:
\begin{equation}
    \text{EMD}(S, T) = \frac{\Sigma^{m,n}_{i=1,j=1}f_{i,j}d_{i,j}}{\Sigma^{m,n}_{i=1,j=1}f_{i,j}}
\end{equation}
where $d_{i,j}=\|s_i - t_j\|$ refers to the $L_1$ distance. The optimal flow $f_{i,j}$ corresponds to the least total work for moving the distance between $S$ and $T$.
Based on EMD, we define {\em conversational similarity} $\text{sim}(S, T)=e^{-\gamma \text{EMD}(S, T)}$, where $\gamma$ is set to be $0.01$.

\section{Data and Experiments}
We used the recipe introduced in \cite{fujita2019end} to generate simulated single-channel meeting mixtures based on two-channel Switchboard-2 (Phase I, II, III), Switchboard Cellular (Part 1 ,2) (SWBD), and the 2004-2008 NIST Speaker Recognition Evaluation (SRE) data sets. All recordings were telephone speech data sampled at 8 kHz. A total of 6,381 speakers was partitioned into 5,743 speakers for training dataset and 638 speakers for testing.
Since these data sets do not contain time annotations, we extracted speech segments using speech  activity  detection  (SAD)  on  the  basis  of  a  time-delay neural network and statistics pooling from a Kaldi recipe.
LibriSpeech (LS) \cite{panayotov2015librispeech} is also used for this mixture generation, comprising 2496 speakers and 970 hours of labeled speech. 
For robustness, we further added noises drawn from 37 background noise recordings in the MUSAN corpus \cite{snyder2015musan} and
sampled from a set of 10,000 room impulse responses (RIRs) from the Simulated Room Impulse Response Database \cite{ko2017study}.

We also used original telephone speech recordings, mapped to a single channel, as the real training and test sets in our experiments.
A set of 64,531 two-speaker recordings were extracted from the recordings of the SWBD and SRE datasets.
We also prepared real conversations from the CALLHOME corpus \cite{callhome}, using only two-speaker conversations.
(SA-EEND requires a fixed maximum number of speakers that, for the present study, was set at two.)
We divided the CALLHOME two-speaker data into two subsets: a fine-tuning set of 155 recordings and a test set of 148 recordings, randomly split according to the recipe in \cite{fujita2019end}.

To investigate the impact of conversational similarity between training and the target CALLHOLME domain, we created different training data sets with the minimum utterance length as an adjustable parameter.
As shown in Table~\ref{tab:data_statistic}, we used a minimum utterance length of either 1.5 seconds (S0, S1, R1) or 0 second (S2, R2).
We also combined simulation (S1, S2) and real (R2) data to construct a larger combined data set (S2+R2, S1+R2) that is larger, yet still similar enough to CALLHOME dataset.
Data statistics for simulation, real and CALLHOME data with their settings are shown in the Table \ref{tab:data_statistic}.
We observe a higher conversational similarity to the CALLHOME test data when a minimum utterance length of 0 second is used (Table \ref{tab:tb_callhome}, columns 4 and 4) highlighting the importance of short turns (e.g., backchannels).
We obtained an upper bound on conversation similarity by comparing the CALLHOME training and test sets, yielding 0.98 for both overlap and silence similarity. 

The acoustic features consisted of 23-dimensional or 80-dimensional log-Mel filter banks with a 25 ms frame length and 10 ms frame shift. SpecAugment \cite{park2019specaugment} was applied with two frequency masks in which each masks at most two consecutive Mel frequency channels ($F=2$), and two time masks where the maximum-size of the mask is set to 1200 consecutive time steps ($T=1200$). For sub-sampling, we used either sub-sampling on stacked frames (used in the original SA-EEND model \cite{fujita2019end}) or convolutional sub-sampling. For sub-sampling on stacked frames, each feature was concatenated with the previous seven frames and subsequent seven frames from the current frame. For convolution sub-sampling, we used two layers of two-dimensional (2D) depth-wise separable convolution \cite{howard2017mobilenets} instead of conventional convolutions due to its efficiency of handling features of long sequence length. Using both methods, we sub-sampled the input features by a factor of ten (corresponding to 100 ms). The parameters for two layers of the 2D convolution are: (1) $\{(3, 3), (7, 7)\}$ for kernels and (2) $\{(2, 1), (5, 1)\}$ for strides with 23-dimensional acoustic features and $\{(2, 2), (5, 2)\}$ with 80-dimensional features. 

For SA-EEND, Transformer-based and Conformer-based EEND, four encoder blocks with 256 attention units containing four heads (P = 4, D = 256, H = 4) are used. We used 1024 and 256 internal units in a position-wise feed-forward layer for Transformer and Conformer, respectively. The effects of different convolution kernel sizes, positional encoding and normalization of residual unit (i.e pre-norm, post-norm) in Conformer blocks are investigated, while keeping the total number of parameters unchanged or less than Transformer blocks. To obtain the optimal configuration for Conformer-based EEND, we ran ablation studies. First, we added relative position encoding to Conformer (in the SA-EEND \cite{fujita2019end}, they did not add positional encoding in Transformer blocks). Next, we replaced pre-norm with post-norm (in the SA-EEND \cite{fujita2019end}, they used pre-norm in Transformer blocks). Finally, we tried different kernel sizes for the convolution module (17, 32) and different dimensions for the FFN (256, 1024). 
We found Conformer blocks with convolution kernel of size 32, the FFN with 256 dimensions, no positional encoding and pre-norm residual units perform the best.

We first trained the model using simulated data (S0, S1, S2), real data (R1, R2), or a combination thereof (S2+R2, S1+R2), with two speakers for 100 epochs, using Adam optimizer with the learning rate scheduler introduced in \cite{vaswani2017attention}, in which warm-up steps were set as 25000. The batch size was 64 for all experiments. 
The average model after 100 epochs was obtained by averaging the model parameters of the last 10 epochs. Finally, the averaged model was fine-tuned using the CALLHOME fine-tuning set, which contains 148 recordings with two speakers. 
We used Adam optimizer for fine-tuning using a fixed learning rate of $10^{-5}$, similar to the original SA-EEND \cite{fujita2019end}. 
Using stochastic gradient descent (SGD) with momentum as optimizer, optimal hyperparameters are explored with a different learning rate (0.01, 0.005, 0.001), momentum (0.9, 0.0) and weight decay (0.0, 0.0001). We evaluated all of our models by their diarization error rates (DER). As is standard, a tolerance of 250\,ms when comparing hypothesized to reference speaker boundaries was used.

\section{Results and Discussion}

\newcommand{\T}{\scriptsize}
\begin{table}[t]
	\caption{Training data characteristics.
	        S0, S1, S2 represent simulated conversations; R1, R2, CH represent real conversations.}
	\label{tab:data_statistic}
	\centering
	\footnotesize
\begin{tabular}{ c | c r | r r r}
	\toprule
    \T Data  & \T Source    & \T Min.   & \T Average & \T Overlap      & \T Total  \\
    \T style &  \T corpora & \T length & \T duration & \T ratio & \T duration \\

    \midrule
    S0 & SWBD+SRE+LS & 1.5s &  166.1s & 48.4 & 9000h \\
    S1 & SWBD + SRE & 1.5s &  88.3s & 34.5 & 2452h  \\
    S2 & SWBD + SRE & 0s &  71.5s & 26.7 & 2482h  \\
    \midrule
    R1 & SWBD + SRE  & 1.5s & 306.1s & 6.5 & 2230h  \\
    R2 & SWBD + SRE  & 0s &  306.5s & 18.3 & 2231h  \\
    \midrule                
    CH  &  CALLHOME & 0s &  74.0s & 14.0 & 3h  \\
    
	\bottomrule
\end{tabular}
\end{table}

\begin{table}[t]
	\caption{DER (\%) on simulated test data.  Best DERs for each condition are highlighted. Baseline DER for an x-vector diarization system is 28.77\% \cite{fujita2019end}.}
	\label{tab:tb_sim}
	\centering
	\footnotesize
\begin{tabular}{ c c |r r r}
	\toprule
	Training & Test & Self-attentive & Transformer- & Conformer- \\
	data &   data & EEND \cite{fujita2019end} & based EEND  & based EEND  \\
	style &  style & (SA-EEND) & (TB-EEND)  & (CB-EEND) \\

	\midrule
    S0 & S0 &  5.09  & 3.44 & \bf 2.73 \\
    
    S1 & S1 &  6.50  & 3.54 & \bf 2.85 \\
    S2 & S2 &  8.76  & 5.94 & \bf 4.60 \\
    \midrule
    
    R2 & S2 & 29.65  & 32.61 & \bf 23.12\\
    \midrule
    S1 + R2 & S1 & 10.35  & 8.60 & \bf 3.28 \\
    
	\bottomrule
\end{tabular}
\end{table}

\begin{table*}[t]
	\caption{DER (\%) when fine-tuning and testing on CALLHOME, and conversational similarities between training and test data.  Hyperparameters were optimized for all systems, but only DERs in italics were thus improved.   Best DERs for each condition are highlighted.  DER for baseline x-vector system is 11.53\% \cite{fujita2019end}. }
	\label{tab:tb_callhome}
	\centering
	\footnotesize
\begin{tabular}{ c  c| r r c |r r r| r r}
	\toprule
	Training &  Test & Overlap & Silence &  Total & Self-Attentive & Transformer- & Conformer- & \multicolumn{2}{c}{Relative improvement} \\
	data &  data & similarity & similarity & training & EEND \cite{fujita2019end} & based EEND  & based EEND  & \multicolumn{2}{c}{with CB-EEND compared to} \\
	style & style & & & duration & (SA-EEND) & (TB-EEND)  & (CB-EEND) & SA-EEND &  TB-EEND\\
 	\midrule
    S0 & CH & 0.31 & 0.26 & 9000 h & 10.60 & \bf 7.63  & \textit{9.35} & 11.8 \% & -22.5 \%  \\
    S1 & CH & 0.50 & 0.59 & 2452 h & \textit{10.52}  & \bf 8.12 &  \textit{9.70} & 7.8 \% & -19.5 \% \\
    S2 & CH & 0.72 & 0.58 & 2482 h & \textit{9.31}   & \bf 8.10 & \textit{8.61} & 7.5 \% & -6.3 \% \\
    \midrule
    R2 & CH & 0.89 & 0.96 & 2231 h & 10.11 & \textit{9.61}  & \bf \textit{7.48} & 26.0 \% & 22.2 \% \\
    \midrule
    S1 + R2 & CH & - & - & 4683 h & \textit{9.01}  & \textit{8.56} & \bf \textit{6.82}  & 24.3 \% & 20.3 \% \\
	\bottomrule
\end{tabular}
\end{table*}

\subsection{Transformer- and Conformer-based EEND}

In reporting results, we refer to the self-attentive EEND \cite{fujita2019end} using standard Transformer architecture as SA-EEND.
We use \textit{Transformer-based EEND}, or TB-EEND, for the system with added SpecAugment \cite{park2019specaugment} and convolutional subsampling.
As shown in Table \ref{tab:tb_sim}, TB-EEND outperforms SA-EEND when simulated data is used for training and testing.
When only simulated data is used (S0, S1, S2) for training, DER improves by up to 45.5\% relative (31.7\% for S0, 45.5\% for S1, 32.2\% for S2). 
When a combination of simulated and real data (S1+R2, the best combination) is used for training, DER improves by 16.9\% relative, still on simulated test data. 
Moving to real data (CALLHOME) for testing and fine-tuning, Table \ref{tab:tb_callhome} shows a similar trend.
In this case, we observe DER improves regardless of the choice of 1st-stage training data.
Therefore, we conclude that SpecAugment and convolutional subsampling combined boost the performance of EEND in general. 

By \textit{Conformer-based EEND}, or CB-EEND, we denote the EEND system including the Conformer encoder (Figure \ref{fig:conformer}). Note that the number of parameters for TB-EEND and CB-EEND is similar (4.4M versus 4.2M parameters).
As shown in Table \ref{tab:tb_sim}, CB-EEND shows a significant DER improvement compared to the original SA-EEND and TB-EEND on simulated test data.
In Table \ref{tab:tb_callhome}, we observe a similar trend for the CALLHOME test set (after fine-tuning on the CALLHOME training portion). CB-EEND outperforms TB-EEND on CALLHOME test data when real data (R2) or a combination of simulation and real data (S1+R2) is used for training (22.2\% and 20.3\% relative improvements, respectively).\footnote{We observed better DER with R2 than with R1, and with S1+R2 than with S2+R2 (results not shown).}
However, CB-EEND underperforms TB-EEND when simulated data (S0, S1, S2) is used for training. Note that CB-EEND still outperforms the x-vector and original SA-EEND systems \cite{fujita2019end} in this case.
 
\subsection{Optimizing Hyperparameters for Fine-Tuning}

As demonstrated in \cite{li2020rethinking}, optimal hyperparameters for fine-tuning are sensitive to the similarity between source and target domain (i.e., train/test mismatch). For instance, it is known that a larger momentum value often results in a better fine-tuning performance for those domains that are dissimilar to the source domain. To closely examine the overall efficacy of hyperparameter optimization (HPO) for fine-tuning, we conducted experiments to further reduce DER using an SGD optimizer with different hyperparameters. 
In Table \ref{tab:tb_callhome} we {\em italicize} those DERs for which HPO resulted in significant gains.
We observed that CB-EEND benefitted from HPO for all training conditions, whereas it had minimal impact on the TB-EEND models when simulated training data is used, even with Adam optimizer.
Specifically, momentum is an important hyperparameter for CB-EEND, as a large momentum value (set to 0.9) helps fine-tuning even when there is a significant mismatch between train and test data, whereas zero momentum works better when the mismatch is negligible. These results are well-aligned with the main findings in \cite{li2020rethinking}, in which various convolution-based architectures are considered. As our Conformer-based EEND contains several convolution components, it is expected that CB-EEND is sensitive to its hyperparameters, which in turn are also sensitive to domain similarity.

\subsection{Effect of Conversational Similarity}

To better understand why CB-EEND improves accuracy on CALLHOME when real data (R2 or S1+R2) is used for training, but hurts accuracy when only simulated data is used for training, we look at overlap- and silence-based similarities between training and test data.
In columns 3 and 4 of Table \ref{tab:tb_callhome}, we observe that all simulated data have a relatively low similarity with CALLHOME test data, and as similarity scores increase (S0, S1, S2), the DER gap between TB-EEND and CB-EEND shrinks. For example, simulated data S0 has low conversational similarity (overlap: 0.31; silence: 0.26), and the DER gap
is 2.12\% absolute.
Simulated data S2, on the other hand, has higher similarity (overlap: 0.72; silence: 0.58), and the DER gap
reduces to 0.51\% absolute. 
Furthermore, we find that for real data from another corpus (R2), similarity scores with CALLHOME test data are still higher 
(overlap: 0.89; silence: 0.96), and CB-EEND now outperforms TB-EEND by 2.13\% absolute.

Though more investigation is clearly needed,
our tentative conclusion is that CB-EEND does relatively better than TB-EEND the more realistic the turn-taking statistics of the training data are with respect to real conversation.
It may be surprising, then, that CB-EEND benefits from mixing simulated data in with 
real training data (i.e., training on S1+R2 is better than on either S1 or R2 alone).
One plausible reason is that the convolutional machinery is good at learning local speaker-change cues from real data, whereas the attention mechanism can learn long-range speaker comparisons even from simulated data. TB-EEND, on the other hand, degrades when data is mixed, compared to training just on simulated data.

\section{Conclusions}

We have demonstrated  that  data  augmentation  and  convolutional  subsampling  layers  enhance  the  original  self-attentive  end-to-end neural diarization (EEND) architecture.  A further improvement is obtained by introducing a Conformer as encoder into the system.
While our Conformer-based EEND model is superior in overall performance, we show that it is more sensitive to larger mismatch between simulated training data and real conversational test data, which we quantify by similarity metrics based on overlap and silence region durations.
Mitigating this sensitivity, we find that the proposed Conformer-based EEND is highly effective when trained on a mixture of simulated and real conversation data, which is not the case for a corresponding Transformer-based system. 
Overall, on two-speaker CALLHOME conversations, we achieve a relative error reduction of 24.3\% over the best baseline SA-EEND training setup,
and of 10.6\% over the best augmented Transformer-based system.

While our results indicate that realistic turn-taking simulation is key for training EEND systems, we have not yet attempted
to improve the simulation recipe from \cite{fujita2019end} to better emulate real speakers in this regard, let alone to ensure the
{\em content} of the speech is realistic in context.
This suggests 
possible future improvements for diarization.

\section{Acknowledgment}
We wish to thank our colleague Oguz Elibol for valuable input.

\normalsize

\bibliographystyle{IEEEtran}
\bibliography{mybib}

\end{document}